\documentclass[preprint]{elsarticle}
\usepackage[utf8]{inputenc}
\usepackage[T1]{fontenc}
\usepackage{graphicx}
\usepackage{grffile}
\usepackage{longtable}
\usepackage{wrapfig}
\usepackage{rotating}
\usepackage[normalem]{ulem}
\usepackage{amsmath}
\usepackage{textcomp}
\usepackage{amssymb}
\usepackage{capt-of}
\usepackage{hyperref}
\usepackage{subfig}
\usepackage{tikz}
\usepackage{booktabs}
\usepackage{commath}
\usepackage{amsmath}
\usepackage{comment}            % Allows to comment out sections of code
\usepackage{todonotes}

\DeclareMathOperator{\E}{E}
\DeclareMathOperator*{\D}{D}
\DeclareMathOperator*{\C}{C}
\DeclareMathOperator*{\rfs}{rfs}
\DeclareMathOperator*{\oo}{o}
\usetikzlibrary{shapes.misc, positioning}

\journal{Neural Networks}
\biboptions{authoryear}

\begin{document}

\begin{frontmatter}

  \title{Arguments for the Unsuitability of Convolutional Neural Networks for Non--Local Tasks}

  \author{Sebastian Stabinger\footnote{sebastian@stabinger.name}, David Peer, Antonio Rodríguez-Sánchez}
  \address{Universität Innsbruck, Technikerstrasse 21a, 6020 Innsbruck, Austria}

  \begin{abstract}
    Convolutional neural networks have established themselves over the
    past years as the state of the art method for image
    classification, and for many datasets, they even surpass humans in
    categorizing images. Unfortunately, the same architectures perform
    much worse when they have to compare parts of an image to each
    other to correctly classify this image.

    Until now, no well-formed theoretical argument has been presented
    to explain this deficiency. In this paper, we will argue that
    convolutional layers are of little use for such problems, since
    comparison tasks are global by nature, but convolutional layers
    are local by design. We will use this insight to reformulate a
    comparison task into a sorting task and use findings on sorting
    networks to propose a lower bound for the number of parameters a
    neural network needs to solve comparison tasks in a generalizable
    way. We will use this lower bound to argue that attention, as well
    as iterative/recurrent processing, is needed to prevent a
    combinatorial explosion.
  \end{abstract}


\end{frontmatter}

% \linenumbers

\newcommand{\etal}{et~al.~}

\section{Introduction}
\label{sec:orgd606471}
Being able to compare objects in a scene and making decisions based on
that information is an essential skill for humans, who can
compare completely novel shapes and objects without being familiar
with them, something which for example \cite{fleuret2011comparing}
were able to show using the SVRT dataset.

Since 2012, deep learning and Convolutional Neural Networks (CNNs)
have emerged as state-of-the-art methods in computer vision. It is
therefore natural to extend the use of such networks to tasks
involving judgments about similarity and identity.

Unfortunately, experiments by us (\cite{stabinger201625}) have shown
that CNNs perform very poorly on classification tasks that require
comparison of shapes. Although multiple authors have confirmed this
problem of CNNs by now (e.g. \cite{ricci2018same} and
\cite{kim2018not}), to our knowledge, no convincing theoretical
arguments on why comparison tasks are so difficult for these
architectures have been proposed in detail.

In this paper, we will try to shed some light on this class of
comparison problems by analyzing one specific task in detail. We will
try to convince the reader of three important aspects regarding
comparison tasks: 1) They are inherently difficult for CNNs, 2)
Attention drastically reduces the size of a network needed to solve
such tasks and 3) Iterative processing further reduces the complexity
of the task considerably. In Section \ref{sec:related}, we will
present the current research on solving comparison tasks using CNNs.
The \emph{identity task}, which we will concentrate our analysis on,
will be presented in Section \ref{sec:identity}. The same section will
also present a theoretical framework to analyze such tasks. Section
\ref{sec:locality} will use this framework to introduce the concept of
the \emph{locality} of a task, which we will use in Section
\ref{sec:application} to argue that the convolutional layers are of
limited usefulness for solving the \emph{identity task}. We will also
show, that the identity task can, in the optimal case, be reduced to
sorting a list of numbers in the fully connected part of a network. In
Section \ref{sec:sortingnetworks}, we will use research on
\emph{sorting networks} to propose a way to generate neural networks
to sort numbers and offer a lower bound on the number of parameters
such a neural network needs. Experiments in Section
\ref{sec:experiments} will fortify some of the theoretical arguments
with practical results. In Section \ref{sec:conclusion} we will
present conclusions and discuss our findings.

\section{Related Work}
\label{sec:related}
\def\imgsize{0.15}
\begin{figure}[tbp]
  \centering
  \subfloat[Class 1]
  {\fbox{\includegraphics[width=\imgsize\textwidth]{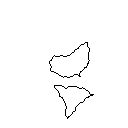}}
    \fbox{\includegraphics[width=\imgsize\textwidth]{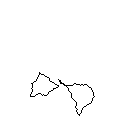}}}
  \hspace{2mm}
  \subfloat[Class 2]
  {\fbox{\includegraphics[width=\imgsize\textwidth]{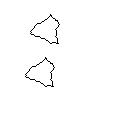}}
    \fbox{\includegraphics[width=\imgsize\textwidth]{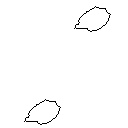}}}
  \hfill
  \caption{Examples of the two classes of problem 1 from the SVRT
    dataset by \cite{fleuret2011comparing}. For class 1 the two shapes
    are different, for class 2 they are identical.}
  \label{fig:prob1_example}
\end{figure}
The SVRT dataset by \cite{fleuret2011comparing} consists of multiple
problems, each built around the classification of abstract images
containing shape outlines. For some problems, it is necessary to
compare the shapes to each other to be able to classify the
image correctly. For example, in problem 1 of the SVRT dataset, two shapes are
visible in each image. The image belongs to class 1 if both shapes are
different, and to class 2 if they are identical (see
\autoref{fig:prob1_example} for a few example images).

\begin{center}\begin{figure}[tbp]
    \centering
    \includegraphics[width=0.4\textwidth]{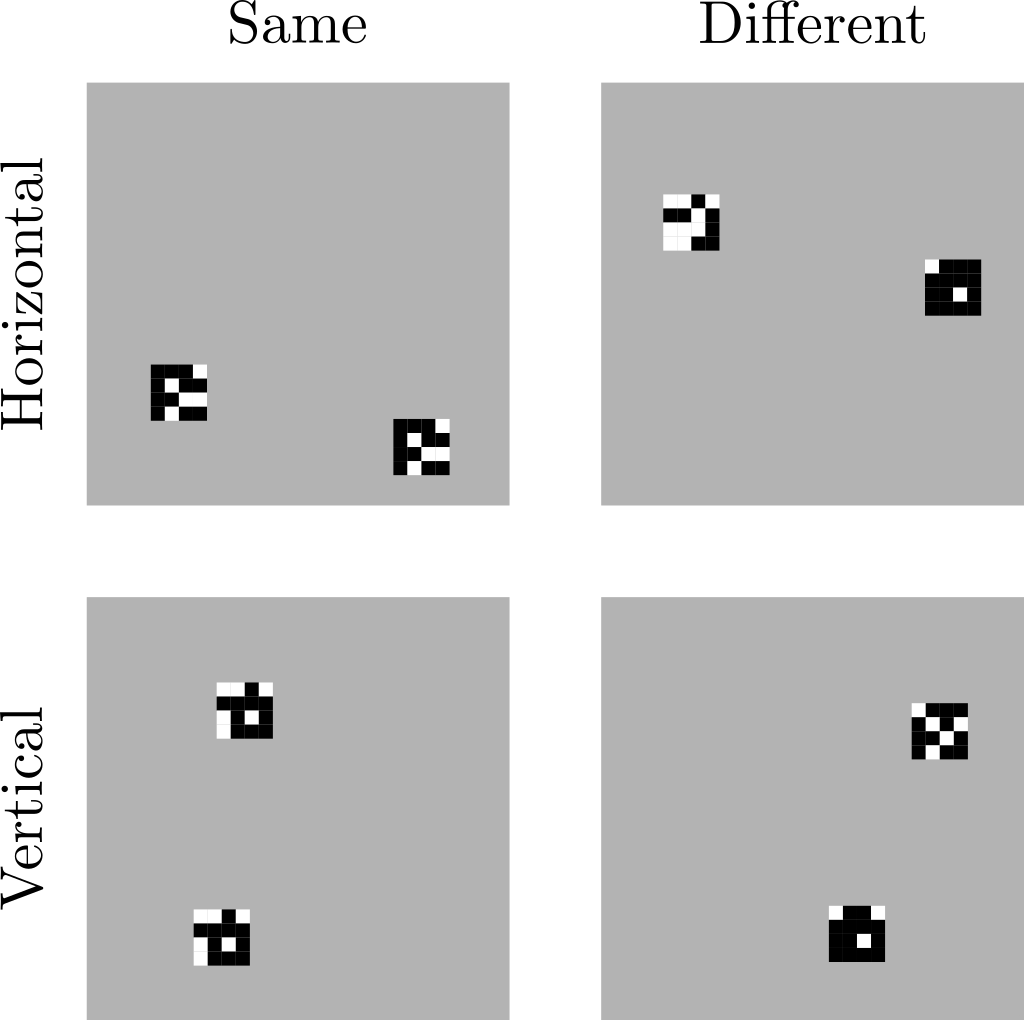}
    \caption{Examples from the PSVRT dataset by \cite{kim2018not} that
      offer two classification tasks with the same images either
      classifying identity of the patches or their orientation to each
      other.\label{fig:psvrt}}
  \end{figure}\end{center}

Using this SVRT dataset, we could show in \cite{stabinger201625}
showed that CNNs struggle with solving tasks that require the
comparison of shapes. \cite{ricci2018same} later came to the same
conclusion and \cite{kim2018not} extended this work by introducing the
PSVRT dataset (see \autoref{fig:psvrt} for examples) using randomly
generated checkerboards instead of shape outlines.

Further research by \cite{messina2019testing} showed that ResNet by
\cite{he2016deep} as well as CorNet-S by \cite{kubilius2018cornet}
can solve some of the previously unsolved
problems of the SVRT dataset, but needed to use 400.000 training
images to do so. \cite{funke2020notorious} finally were able to
achieve accuracies above 90\% on all problems of the SVRT dataset
using a ResNet architecture with just 28.000 images.

The fact that the SVRT dataset has been solved by
\cite{funke2020notorious} probably has more to do with shortcomings of
the dataset and less with neural networks being able to solve
the underlying comparison problem. Our experiments with a seemingly
easier dataset in Section \ref{sec:attention} seem to support this
hypothesis. In our opinion, the SVRT dataset has two shortcomings:

\emph{First}, the images for the different problems also have
different complexity (e.g. the number, size and distribution of the
shapes). Therefore, it is hard to judge whether a system struggles
because the actual problem is harder to solve, or because the images
have higher variability than images for other problems (e.g. there are
more shapes, the size of the shapes is different, the positions of the
shapes are more variable, ...). \cite{kim2018not} created the PSVRT
(see \autoref{fig:psvrt}) dataset to take image complexity out of the
equation by using the same images for two different tasks (spatial
orientation or identity). The authors used PSVRT to confirm that tasks
involving comparison are truly more difficult to learn than those
involving other relations (like spatial relations in the PSVRT case).

\emph{Second}, even an approximate comparison of the shapes of the
SVRT dataset is sufficient to claim identity. Since the shapes are
generated entirely at random, different shapes, in most cases, do not
resemble each other even on a coarse level. Thus, two shapes that
roughly look-alike will be identical with quite a high probability.
Because of this, even a small number of training images might be
sufficient to have seen all approximate shapes and the separation
between training and testing set can not be ensured anymore. This
means that it is hard to judge whether a system has learned a task or
has just memorized the training samples.

Humans, as well as more classical machine learning approaches like
support vector machines with handcrafted features, do not show this
systematically lower performance on tasks involving comparison, which
was shown by Fleuret \etal in their original paper.

\section{The Identity Task \label{sec:identity}}
\label{sec:orgb4f089b}
\begin{figure}[tbp]
  \centering
  \subfloat[Class one]
  {\includegraphics[width=0.25\textwidth]{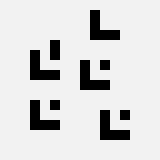}}
  \hspace{2mm}
  \subfloat[Class two]
  {\includegraphics[width=0.25\textwidth]{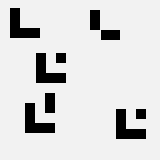}}
  \hfill
  \caption{ Example images from the identity task, rescaled for better
    visibility. Since in a), the same patch appears three times it is
    an example of class one. For b), no patch occurs at least three
    times, therefore it is an example of class two.}
  \label{fig:ex}
\end{figure}

Because of these shortcomings, we will use a simplified task in this
paper that was influenced by the identity task of the PSVRT dataset by
\cite{kim2018not}. The task consists of images \(\mathcal{I}\) with
dimension \(N \times N\) where each pixel can have one of two possible
states \(\in \{0, 1\}\). Each image contains a randomly chosen amount
\(c \geq 3\) of non-overlapping, randomly positioned patches
\(\mathcal{P}\) of size \(n \times n\), where \(n \ll N\). The patches are
filled with random pixels, according to a Bernoulli distribution with
\(p=0.5\) and \(\mathcal{P}_{xy} \in \{ 0, 1 \}\). Each image is
categorized into one of two classes. An image is of \emph{class one}
if at least half of the patches are identical, and \emph{class two}
otherwise. Two patches \(\mathcal{P}^i\) and \(\mathcal{P}^j\) are
identical iff
\(\forall x,y \in \{1,...,n\} : \mathcal{P}^i_{xy} = \mathcal{P}^j_{xy}\). The
goal of this task is to decide which class a given image belongs to,
given \(M\) image/class pairs for training. We call this
classification task and the accompanying dataset the \emph{identity
  task}. Example images from this dataset can be seen in
\autoref{fig:ex}.

\subsection{Determining the Usefulness of Convolutional Layers}
\label{sec:orgcbee804}

In this section, we will present arguments why convolutional layers
are of limited usefulness for solving the identity task. As an
abstraction of our actual task, let us assume that some system has to
compare two image patches and decide whether they are identical or
not.

We assume that all patches that should be considered identical are
assigned one unique symbol by a function \(\E(p)\) where \(p\) is the
patch to be mapped to a symbol. We will call this function the
\emph{encoder}. The set of all possible symbols for a task will be named
\(S\) so that \(\forall p \E(p) \in S\). One symbol therefore encodes
\(\log_2 |S|\) bit of information, where \(|S|\) signifies the number of
symbols in \(S\).

The task of comparing two patches for identity can now be solved by a
decision function \(\D(s_1, s_2)\) taking in two symbols. This function
detects whether \(s_1\) and \(s_2\) are identical or not (see
\autoref{eq:decision}), and we will call it the \emph{decision--maker}.

\begin{equation}
  \label{eq:decision}
  \begin{gathered}
    \D(s_1, s_2) =
    \begin{cases}
      \text{same} \text{ iff } s_1 = s_2\\
      \text{different otherwise}
    \end{cases} \\
  \end{gathered}
\end{equation}
Deciding whether the patches \(p_1\) and \(p_2\) are identical
therefore is reduced to the question of whether the result from
\(\D(\E(p_1), \E(p_2))\) is "same" or "different".

In our identity task, the goal is not to compare two patches for
identity, but to compare \(c\) patches. A decision--maker, processing
all patches in one step, therefore, would need \(c\) symbols as input
and therefore \(c\log_2 |S|\) bit of information from the encoders. We
will now investigate the following question: Can we somehow reduce the
amount of information the decision--maker needs to solve a given task
by some form of hierarchical preprocessing?
\subsubsection{Hierarchical Preprocessing}
\begin{center}\begin{figure}[tbp]
    \centering
    \includegraphics[width=0.5\textwidth]{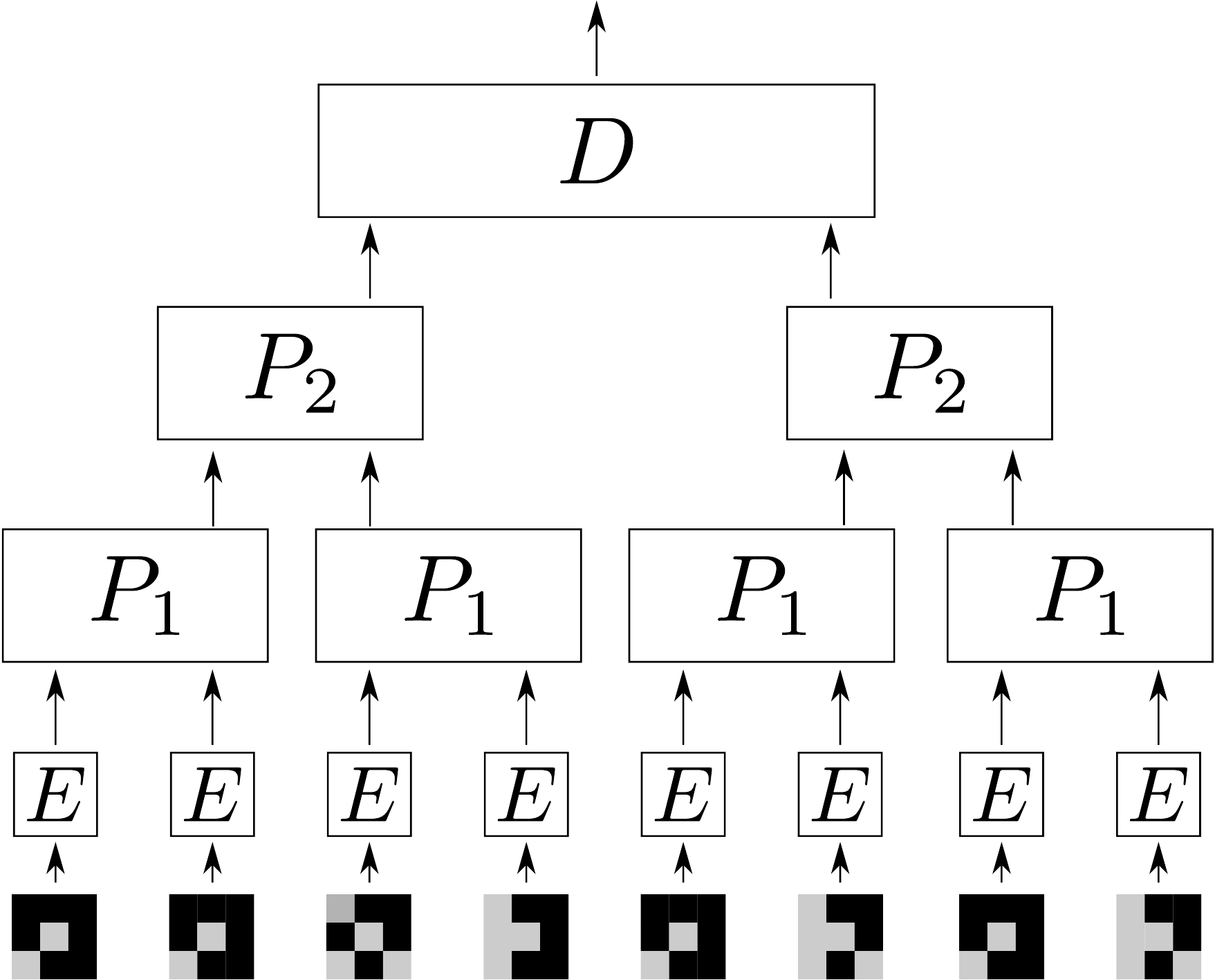}
    \caption{Example of a hierarchical arrangement of encoders \(E\),
      preprocessors \(P_1\) and \(P_2\), and a decision--maker \(D\)
      for the presented identity task \label{fig:hierrec}}
  \end{figure}\end{center}

We define a \emph{preprocessor} \(P_i(s_1, s_2)\) as a function that
takes two symbols from a set of symbols \(S_{i-1}\) as arguments and
returns a single symbol from a new set of symbols \(S_i\). A
preprocessor, therefore, performs lossless compression of the received
information with respect to the given task. How much a preprocessor
can compress the incoming information depends on the task that has to
be solved. The ratio between the amount of information going into a
preprocessor and the amount of information being passed along to
higher-level preprocessors will be called the \emph{compression
  factor}, defined as
$$\C(P_i) = \frac{2\log_2|S_{i-1}|}{\log_2|S_i|}$$
This compression factor directly correlates to the amount of
processing that can be performed in the preprocessor \(P_i\). A high
compression factor indicates that a lot of the processing can be done
in \(P_i\), and a compression factor of \(1\) indicates that no
processing can occur in \(P_i\) and all the incoming information has
to be passed along unchanged. Preprocessors can be stacked in a
hierarchical arrangement (see \autoref{fig:hierrec}).

\section{Locality}
\label{sec:locality}
Given a preprocessor \(P_i\), we define the \emph{receptive field
  size} \(\rfs(P_i)\) as the number of inputs its output, directly or
indirectly, depends on. E.g. for the hierarchical structure from
\autoref{fig:hierrec}, the first layer of preprocessors (\(P_1\)) will
have a receptive field size of 2, the next layer will have a receptive
field size of 4 and so forth (i.e. \(\rfs(P_i) = 2^i\))

We will now investigate how the minimum amount of information that is
needed to fully represent the content of a specific receptive field
changes with respect to its size. This minimum amount of information
is equivalent to the number of output symbols a preprocessor in our
hierarchical processing scheme needs.

If the minimal amount of information needed to represent a receptive
field does not depend on the receptive field size, we have
\emph{perfect locality} (i.e. all the processing can be done on a
local scope). If the amount of information grows directly proportional
to the receptive field size we have \emph{no locality} at all (i.e.
all processing has to be done on a global scope by the
decision-maker).

Therefore, we define the locality \(L\) of a task as the limit of the
compression factor of our preprocessors while going up the hierarchy
and therefore simultaneously increasing the receptive field size of
the preprocessors (we assume that the receptive field size and the
depth of the hierarchy are not limited):
$$L = \lim_{i \rightarrow \infty}\C(P_i) - 1$$

Informally, taking the identity task as an example, this can be seen
as having an infinite number of patches to analyze. Therefore, the
hierarchy of \autoref{fig:hierrec} would have to have infinite depth,
and the locality is the compression factor the preprocessors approach
while going up this infinite hierarchy, minus one.

A locality of \(0\) indicates that no local processing can be
performed if the receptive field becomes very big (i.e. the
preprocessor is not able to compress anything). A locality \(L < 0\)
can not occur since it is always possible to pass along the full
incoming information. A locality of \(1\) means that all processing
can be done on a local level, i.e. the preprocessor can compress two
symbols from \(S_{i-1}\) to one symbol from \(S_i\) and the amount of
information needed to represent one symbol from \(S_{i-1}\) and
\(S_i\) is the same (\(|S_{i-1}| = |S_i|\)). In other words, the
amount of information needed to represent a receptive field does not
depend on its size. A locality \(L\) between \(0\) and \(1\) indicates
that a varying degree of processing can be performed on a local level.
\begin{description}
\item[{Hypothesis:}] Problems exhibiting low locality are ill fitted to be
  solved by Convolutional Neural Networks (CNNs)
\end{description}
Intuitively, CNNs are ill suited to solve non-local tasks, since the
convolutional part of the network is local by design.
\subsubsection{The Identity Task as an Example}
\label{sec:org9523ba5}
As previously described, we assume that \(c\) patches of size
\(n \times n\), consisting of black and white pixels, are given, and the
task is to decide whether at least half of the patches are identical
or not. Since two patches are only considered equal if they contain
the same pattern, the encoder needs to use \(2^{n^2}\) symbols in
\(S_0\) to preserve comparability. Thus, one such symbol encodes
\(n^2\) bit of information.

The preprocessors of the first layer \(P_1\) each get \(2n^2\) bit of
information as input, assuming that each possible pattern can occur
with the same probability. No matter whether the two incoming symbols
are the same or different, the output has to encode all possible
combinations of incoming symbols since no decision about which patch
has to be available for comparison at a later stage can be made at
this point. Since the order of symbols is not relevant (i.e.
\(P_1(s_1, s_2) = P_1(s_2, s_1)\)), the preprocessor has to use
\(|S_0|^2 - {|S_0| \choose 2}\) symbols for \(S_1\).

Intuitively, one would assume that at least some information can be
left out in cases where a preprocessor receives the same symbol twice
(meaning that the two patches are identical), but this is not the
case. The information that the same patch was received twice needs one
symbol as a representation, the same way that any other combination of
received patches needs exactly one symbol to represent.

This gives us the following compression factor for \(P_1\):
\begin{equation}
  C(P_1) =
  \frac{2\log_2|S_0|}{\log_2\left(|S_0|^2 - {|S_0| \choose 2}\right)}
\end{equation}
The preprocessors of the next layer \(P_2\) again have to use one
symbol in \(S_2\) for all possible pairs of symbols from \(S_1\),
ignoring the order of symbols. This procedure does not change for any
of the following layers of preprocessors, and we can define a general
formula for the compression factor of any preprocessor as
\begin{equation}
  C(P_i) =
  \frac{2\log_2|S_{i-1}|}{\log_2\left(|S_{i-1}|^2 - {|S_{i-1}| \choose
        2}\right)}
\end{equation}
For better readability we define \(s = |S_{i-1}|\). Since the number of
symbols in \(S_i\) monotonically increases while going up the hierarchy,
we can rewrite the locality of this task as follows:
\begin{equation}
  L = \lim_{s
    \rightarrow \infty}\frac{2\log_2 s}{\log_2\left(s^2 - {s \choose
        2}\right)} - 1 = \lim_{s \rightarrow \infty}\frac{2\log_2
    s}{\log_2\left(s^2 - \frac{s^2 - s}{2}\right)} -1 =
\end{equation}
\begin{equation}
  2\lim_{s \rightarrow \infty}\frac{\od{}{s}\log_2
    s}{\od{}{s}\log_2\left(s^2 - \frac{s^2 - s}{2}\right)} - 1 = 2\lim_{s
    \rightarrow \infty}\frac{s+1}{2s+1} - 1 = 0
\end{equation}
With a locality of 0, we can see that processing can only happen once
the system has reached a global receptive field.

\section{Application to the Identity Task \label{sec:application}}
\label{sec:orge009ecc}
We will now see how the theoretical architecture of encoder,
preprocessor and decision--maker can be translated to the different
parts of a Convolutional Neural Network (CNN). Commonly, a CNN for
classification is constructed from two main parts: A varying amount of
convolutional and pooling layers extracting feature descriptors with
increasingly global scope and a part of fully connected layers
generally considered to use those features for the actual
classification.

The symbols the encoder is extracting can be interpreted as the
activations of the neurons of some of the first layers in the CNN
where the receptive fields of the neurons are still relatively small.
The following convolutional and pooling layers can be interpreted as
the preprocessors that compress and extract information as the
receptive field becomes more extensive, either through pooling or
through the stride of the convolutions. The following fully connected
layers have a global receptive field and can be seen as the
decision--maker, taking the extracted information from the
convolutional layers and calculating the class probabilities.

As we have shown in Section \ref{sec:locality}, the only operation
that can be performed on a local scope is extracting the information
contained in the patches, and forwarding this information to the
decision--maker. Therefore, the best the convolution and pooling part
of the network can do is to find an efficient, lossless encoding of
the information contained in each patch. If we allow for arbitrary
precision real numbers, a convolution can in principle completely
encode the content of a patch in a single real-valued output. Those
real-valued outputs can be interpreted as the symbols forwarded to the
decision--maker.

\begin{center}\begin{figure}[tbp]
    \centering
    \includegraphics[width=0.25\textwidth]{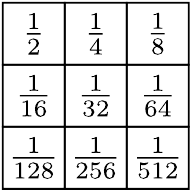}
    \caption{Example of a kernel that is able to fully encode an
      incoming binary patch of size \(3\times3\) in its
      output. \label{fig:weights}}
  \end{figure}\end{center}

A kernel of size \(n \times n\) that is capable of capturing all
information of a patch with binary pixels (i.e. black or white) in a
single real number can be constructed by setting the \(n^2\) weights
of the kernel to \(\forall i \in {1,...,n^2} : w_i = \frac{1}{2^i}\). This
ensures that each bit of the incoming patch is represented by exactly
one bit of the fraction bits of the resulting floating-point number.
See \autoref{fig:weights} for an example of a $3\times3$ kernel constructed
in this manner.

There are two possibilities when extracting the patches in the
convolutional layers. If some attentional mechanism allows the
convolutional layers to extract exactly one number for each patch, we
end up with \(c\) (the number of patches) numbers that need to be
compared. Since we have to design our network in a way so that the
worst case can also be handled, we have to assume the maximal number
of patches that are possible. This means that we have to allow for
\(\left(N/n\right)^2\) numbers to be compared where \(N\) is the width
and height of the image in pixels and \(n\) is the width and height of
the patch in pixels. If no attentional mechanism is present, the
convolutional layers will have to generate one number for each
possible patch position, resulting in \(\left(N-1\right)^2\) numbers
that have to be processed.

In reality, the situation is even worse since the precision of
floating-point numbers is limited. Using a 32-bit floating-point
number (the most common size used for neural networks) only 24 bit can
be reliably encoded with the presented method. Patches above a size of
\(4 \times 4\) would therefore have to be encoded by multiple numbers.

After the convolution and pooling part of the network, in the most
favourable case, we would have several inputs to our fully connected
layers, and each input encodes the information for each patch (in the
case with attention) or each possible patch position (in the case
without attention). The most efficient way of detecting the amount of
identical numbers in a list is to sort the list first. Once the list
is sorted, identical numbers can be identified by merely checking
neighbouring numbers for equality. In the experimental section, we
will show that this also holds in practice using neural networks.

The question now is, how a neural network can sort numbers and how big
it has to be to do so. The next section will start with sorting
networks to estimate a lower bound on the number of parameters needed
to achieve sorting using a neural network under the assumption that it
is constructed from fully connected layers.

\section{Sorting Networks \label{sec:sortingnetworks}}
\label{sec:org8d10102}
\begin{figure}[]
  \centering
  \begin{tikzpicture}
    \tikzstyle{startnode}=[left, scale=1]
    \tikzstyle{wire}=[above right, scale=0.6]
    \tikzstyle{innernode}=[above right, scale=0.7]
    \tikzstyle{endnode}=[right, scale=1]
    \tikzstyle{layernode}=[scale=0.9]

    \draw[dashed] (0.5,0) rectangle (2.7,4.5);
    \node[layernode] at (1.6,0.3) {$L_1$};
    \draw[dashed] (3.5,0) rectangle (4.7,4.5);
    \node[layernode] at (4.1,0.3) {$L_2$};
    \draw[dashed] (5.5,0) rectangle (6.7,4.5);
    \node[layernode] at (6.1,0.3) {$L_3$};

    % \fill [gray!15] (1.5,1.5) -- (2.5,1.5) -- (2.5,2.5) -- (1.5,2.5) -- cycle;
    \foreach \a in {1,...,4}
    \draw[thick] (0,\a) -- ++(7,0);
    \foreach \x in {{1,2},{1,4},{2,1},{2,3},{4,1},{4,2},{4,3},{4,4},{6,2},{6,3}}
    \filldraw (\x) circle (2pt);

    \draw[thick] (1,2) -- (1,4);
    \draw[thick] (2,1) -- (2,3);
    \draw[thick] (4,1) -- (4,2);
    \draw[thick] (4,3) -- (4,4);
    \draw[thick] (6,2) -- (6,3);

    \node[startnode] at (0,4) {0.3};
    \node[startnode] at (0,3) {0.1};
    \node[startnode] at (0,2) {0.6};
    \node[startnode] at (0,1) {0.2};

    \node[wire] at (0,4) {$W_1$};
    \node[wire] at (0,3) {$W_2$};
    \node[wire] at (0,2) {$W_3$};
    \node[wire] at (0,1) {$W_4$};

    \node[innernode] at (1,4) {\textbf{0.6}};
    \node[innernode] at (1,3) {0.1};
    \node[innernode] at (1,2) {\textbf{0.3}};
    \node[innernode] at (1,1) {0.2};

    \node[innernode] at (2,4) {0.6};
    \node[innernode] at (2,3) {\textbf{0.2}};
    \node[innernode] at (2,2) {0.3};
    \node[innernode] at (2,1) {\textbf{0.1}};

    \node[innernode] at (4,4) {0.6};
    \node[innernode] at (4,3) {0.2};
    \node[innernode] at (4,2) {0.3};
    \node[innernode] at (4,1) {0.1};

    \node[innernode] at (6,4) {0.6};
    \node[innernode] at (6,3) {\textbf{0.3}};
    \node[innernode] at (6,2) {\textbf{0.2}};
    \node[innernode] at (6,1) {0.1};

    \node[endnode] at (7,4) {0.6};
    \node[endnode] at (7,3) {0.3};
    \node[endnode] at (7,2) {0.2};
    \node[endnode] at (7,1) {0.1};

  \end{tikzpicture}
  \caption{Example of a sorting network that is able to sort four real numbers in
    descending order. The horizontal lines represent the wires, the
    vertical lines and black dots represent the comparators. Values
    flow along the wires from left to right. Swapped values are
    highlighted and the three layers of the network are labeled $L_1$ through $L_3$.\label{fig:sortingnetwork} }
\end{figure}
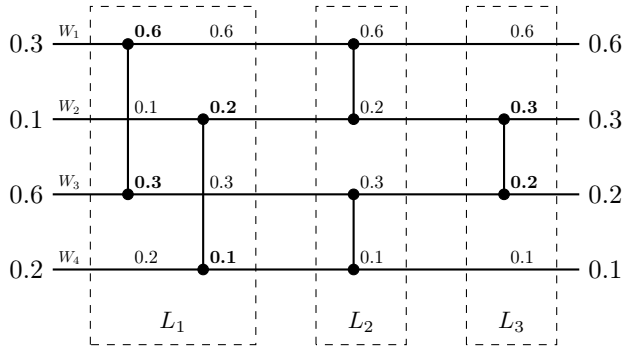

A sorting network by \cite{oconnor1962nuline} consists of \emph{wires}
and \emph{comparators}. Wires "transport" comparable values (e.g. real
numbers in our case). Pairs of wires can be connected by comparators
that swap the values transported on the wires if they are not already
in the correct order. Multiple comparators can swap values in parallel
as long as each wire is only connected to one comparator. We will call
such a parallel evaluation of comparators a \emph{layer}. A
\emph{sorting network} is a fixed arrangement of comparators and wires
so that any combination of possible values sent along the wires is
sorted after passing all comparators. \autoref{fig:sortingnetwork}
shows a sorting network in operation. Sorting networks take the task
of sorting - which is usually perceived as an iterative process - and
converts it into a highly parallel, purely feed-forward problem.

The number of layers of a sorting network is also called its depth.
Extensive research exists on the theoretical properties of such
sorting networks, and lower bounds on their depth have been published.
There exists an information-theoretical lower bound of \(\log_2n\)
regarding the depth of a sorting network to sort \(n\) numbers and
Kahale et al. \cite{kahale1995lower} were able to tighten that lower
bound to \((c - \oo(1))\log_2n\) with \(c \approx 3.27\).

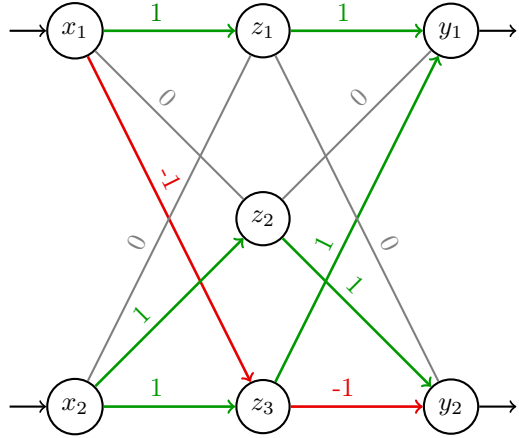
\begin{figure}[]
  \centering
  \begin{tikzpicture}[thick, node distance=2.5cm]
    \tikzstyle{neuron}=[draw=black, shape=circle]
    \tikzstyle{empty}=[node distance=1cm]
    \tikzstyle{vert}=[pos=0.4, above, sloped]
    \tikzstyle{posedge} = [green!60!black, line width=1pt, ->]
    \tikzstyle{negedge} = [red!90!black, line width=1pt, ->]
    \node[neuron] (z1)                    {$z_1$};
    \node[neuron] (z2)    [below of=z1]   {$z_2$};
    \node[neuron] (z3)    [below of=z2]   {$z_3$};
    \node[neuron] (x1)    [left of=z1]    {$x_1$};
    \node[empty]  (X1)    [left of=x1]    {};
    \node[neuron] (x2)    [left of=z3]    {$x_2$};
    \node[empty]  (X2)    [left of=x2]    {};
    \node[neuron] (y1)    [right of=z1]   {$y_1$};
    \node[empty]  (Y1)    [right of=y1]   {};
    \node[neuron] (y2)    [right of=z3]   {$y_2$};
    \node[empty]  (Y2)    [right of=y2]   {};

    \path
    (X1) edge [->]  node {}  (x1)
    (X2) edge [->]  node {}  (x2)
    (y1) edge [->]  node {}  (Y1)
    (y2) edge [->]  node {}  (Y2)
    (x1) edge [posedge]       node [vert, green!50!black] 	{1}  (z1)
    (x1) edge [black!50]      node [vert, black!50] 		{0}  (z2)
    (x1) edge [negedge]       node [vert, red!90!black] 	{-1} (z3)
    (x2) edge [black!50]      node [vert, black!50] 		{0}  (z1)
    (x2) edge [posedge]       node [vert, green!60!black] 	{1}  (z2)
    (x2) edge [posedge]       node [vert, green!60!black] 	{1} (z3)
    (z1) edge [posedge]       node [vert, green!60!black] 	{1}  (y1)
    (y1) edge [black!50]      node [vert, black!50] 		{0}  (z2)
    (z3) edge [posedge]       node [vert, green!60!black] 	{1} (y1)
    (y2) edge [black!50]      node [vert, black!50] 		{0}  (z1)
    (z2) edge [posedge]       node [vert, green!60!black] 	{1}  (y2)
    (z3) edge [negedge]       node [vert, red!90!black] 	{-1} (y2);
  \end{tikzpicture}
  \caption{One possible, minimal implementation of a comparator as a neural network.\label{fig:comparator} }
\end{figure}

Since we want to transfer the knowledge on how to construct a sorting
network to neural networks, we first have to find a way to implement a
comparator using a neural network. We assume the neural network has
two inputs \(x_1\) and \(x_2\) in the range \([0,1]\) and two outputs
\(y_1\) and \(y_2\). We expect the following behavior: \(y_1 = x_1\)
and \(y_2 = x_2\) in case \(x_1 \geq x_2\) and \(y_1 = x_2\),
\(y_2 = x_1\) in case of \(x_1 < x_2\). \autoref{fig:comparator} shows
a minimal implementation of such a comparator neural network. The
implementation assumes \(\max(0, x)\) (a rectified linear unit) as the
activation function of the neurons and biases are not needed.
The following equations describe the hidden neurons:
$$z_1 = \max(0, x_1) \rightarrow z_1 = x_1$$
$$z_2 = \max(0, x_2) \rightarrow z_2 = x_2$$
$$z_3 = \max(0, x_2 - x_1)$$
So if \(x_1 \geq x_2\) then \(z_3 = 0\) otherwise \(z_3 = x_2 - x_1\). This
means we can calculate \(y_1\) and \(y_2\) by
$$y_1 = z_1 + z_3 = x_1 + \max(0, x_2 - x_1)$$
$$y_2 = z_1 - z_3 = x_2 - \max(0, x_2 - x_1)$$
In case of \(x_1 \geq x_2\)
$$y_1 = x_1 + 0 = x_1$$
$$y_2 = x_2 - 0 = x_2$$
In case of \(x_1 < x_2\)
$$y_1 = x_1 + x_2 - x_1 = x_2$$
$$y_2 = x_2 - x_2 + x_1 = x_1$$
Which is the expected behavior of a comparator.

\subsection{Constructing Neural Networks from Sorting Networks}
\label{sec:orgbc7cb30}
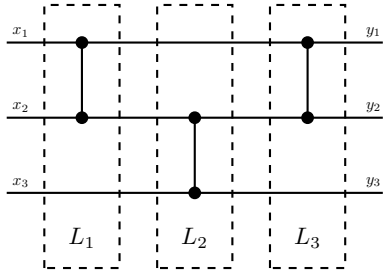
\begin{figure}[tbp]
  \centering
  \begin{tikzpicture}[thick, node distance=2.5cm]
    \usetikzlibrary{shapes.misc, positioning}
    \tikzstyle{startnode}=[left, scale=1]
    \tikzstyle{wire}=[above right, scale=0.6]
    \tikzstyle{innernode}=[above right, scale=0.7]
    \tikzstyle{endnode}=[right, scale=1]
    \tikzstyle{layernode}=[scale=0.9]

    \draw[dashed] (0.5,1) rectangle (1.5,4.5);
    \node[layernode] at (1,1.4) {$L_1$};
    \draw[dashed] (2,1) rectangle (3,4.5);
    \node[layernode] at (2.5,1.4) {$L_2$};
    \draw[dashed] (3.5,1) rectangle (4.5,4.5);
    \node[layernode] at (4,1.4) {$L_3$};

    \foreach \a in {2,...,4}
    \draw[thick] (0,\a) -- ++(5,0);
    \foreach \x in {{1,3},{1,4}, {2.5,2}, {2.5,3}, {4,3}, {4,4}}
    \filldraw (\x) circle (2pt);

    \draw[thick] (1,3) -- (1,4);
    \draw[thick] (2.5,2) -- (2.5,3);
    \draw[thick] (4,3) -- (4,4);

    \node[wire] at (0,4) {$x_1$};
    \node[wire] at (0,3) {$x_2$};
    \node[wire] at (0,2) {$x_3$};
    \node[wire] at (4.7,4) {$y_1$};
    \node[wire] at (4.7,3) {$y_2$};
    \node[wire] at (4.7,2) {$y_3$};
  \end{tikzpicture}
  \caption{Example of an optimal sorting network for three numbers.\label{fig:three_sort_net} }
\end{figure}

\begin{figure}[tbp]
  \centering
  \begin{tikzpicture}[node distance=2.5cm, scale=0.8]
    \tikzstyle{neuron}=[draw=black, shape=circle]
    \tikzstyle{empty}=[node distance=1cm]
    \tikzstyle{vert}=[pos=0.27, above, sloped, inner sep=1pt]
    \tikzstyle{postxt}=[vert, green!50!black]
    \tikzstyle{negtxt}=[vert, red!90!black]
    \tikzstyle{posedge} = [green!60!black, line width=1pt, ->]
    \tikzstyle{negedge} = [red!90!black, line width=1pt, ->]
    \tikzstyle{layernode}=[scale=0.9]

    \node[neuron] at (2, -0.0) (n_1_1) {};
    \node[empty]  (x1)    [left of=n_1_1]    {x1};
    \path (x1) edge [->]  node {}  (n_1_1);
    \node[neuron] at (2, -1.14) (n_1_2) {};
    \node[empty]  (x2)    [left of=n_1_2]    {x2};
    \path (x2) edge [->]  node {}  (n_1_2);
    \node[neuron] at (2, -2.28) (n_1_3) {};
    \node[empty]  (x3)    [left of=n_1_3]    {x3};
    \path (x3) edge [->]  node {}  (n_1_3);
    \node[neuron] at (4, -0.0) (n_2_1) {};
    \node[neuron] at (4, -0.7599999999999999) (n_2_2) {};
    \node[neuron] at (4, -1.5199999999999998) (n_2_3) {};
    \node[neuron] at (4, -2.28) (n_2_4) {};
    \node[neuron] at (6, -0.0) (n_3_1) {};
    \node[neuron] at (6, -1.14) (n_3_2) {};
    \node[neuron] at (6, -2.28) (n_3_3) {};
    \node[neuron] at (8, -0.0) (n_4_1) {};
    \node[neuron] at (8, -0.7599999999999999) (n_4_2) {};
    \node[neuron] at (8, -1.5199999999999998) (n_4_3) {};
    \node[neuron] at (8, -2.28) (n_4_4) {};
    \node[neuron] at (10, -0.0) (n_5_1) {};
    \node[neuron] at (10, -1.14) (n_5_2) {};
    \node[neuron] at (10, -2.28) (n_5_3) {};
    \node[neuron] at (12, -0.0) (n_6_1) {};
    \node[neuron] at (12, -0.7599999999999999) (n_6_2) {};
    \node[neuron] at (12, -1.5199999999999998) (n_6_3) {};
    \node[neuron] at (12, -2.28) (n_6_4) {};
    \node[neuron] at (14, -0.0) (n_7_1) {};
    \node[empty]  (y1)    [right of=n_7_1]    {y1};
    \path (n_7_1) edge [->]  node {} (y1);
    \node[neuron] at (14, -1.14) (n_7_2) {};
    \node[empty]  (y2)    [right of=n_7_2]    {y2};
    \path (n_7_2) edge [->]  node {} (y2);
    \node[neuron] at (14, -2.28) (n_7_3) {};
    \node[empty]  (y3)    [right of=n_7_3]    {y3};
    \path (n_7_3) edge [->]  node {} (y3);
    \path
    % Layer 1 pos
    (n_1_1) edge [posedge]       node [postxt] 	{\tiny 1}  (n_2_1)
    (n_1_2) edge [posedge]       node [postxt] 	{\tiny 1}  (n_2_2)
    (n_1_2) edge [posedge]       node [postxt] 	{\tiny 1}  (n_2_3)
    (n_1_3) edge [posedge]       node [postxt] 	{\tiny 1}  (n_2_4)
    % Layer 1 neg
    (n_1_1) edge [negedge]       node [negtxt] 	{\tiny -1}  (n_2_3)
    % Layer 2
    (n_2_1) edge [posedge]       node [postxt] 	{\tiny 1}  (n_3_1)
    (n_2_2) edge [posedge]       node [postxt] 	{\tiny 1}  (n_3_2)
    (n_2_3) edge [posedge]       node [postxt] 	{\tiny 1}  (n_3_1)
    (n_2_4) edge [posedge]       node [postxt] 	{\tiny 1}  (n_3_3)
    % Layer 2 neg
    (n_2_3) edge [negedge]       node [negtxt] 	{\tiny -1}  (n_3_2)
    % Layer 3
    (n_3_1) edge [posedge]       node [postxt] 	{\tiny 1}  (n_4_1)
    (n_3_2) edge [posedge]       node [postxt] 	{\tiny 1}  (n_4_2)
    (n_3_3) edge [posedge]       node [postxt] 	{\tiny 1}  (n_4_3)
    (n_3_3) edge [posedge]       node [postxt] 	{\tiny 1}  (n_4_4)
    % Layer 3 neg
    (n_3_2) edge [negedge]       node [negtxt] 	{\tiny -1}  (n_4_4)
    % Layer 4
    (n_4_1) edge [posedge]       node [postxt] 	{\tiny 1}  (n_5_1)
    (n_4_1) edge [posedge]       node [postxt] 	{\tiny 1}  (n_5_1)
    (n_4_2) edge [posedge]       node [postxt] 	{\tiny 1}  (n_5_2)
    (n_4_3) edge [posedge]       node [postxt] 	{\tiny 1}  (n_5_3)
    (n_4_4) edge [posedge]       node [postxt] 	{\tiny 1}  (n_5_2)
    % Layer 4 neg
    (n_4_4) edge [negedge]       node [negtxt] 	{\tiny -1}  (n_5_3)
    % Layer 5
    (n_5_1) edge [posedge]       node [postxt] 	{\tiny 1}  (n_6_1)
    (n_5_2) edge [posedge]       node [postxt] 	{\tiny 1}  (n_6_2)
    (n_5_2) edge [posedge]       node [postxt] 	{\tiny 1}  (n_6_3)
    (n_5_3) edge [posedge]       node [postxt] 	{\tiny 1}  (n_6_4)
    % Layer 5 neg
    (n_5_1) edge [negedge]       node [negtxt] 	{\tiny -1}  (n_6_3)
    % Layer 6
    (n_6_1) edge [posedge]       node [postxt] 	{\tiny 1}  (n_7_1)
    (n_6_2) edge [posedge]       node [postxt] 	{\tiny 1}  (n_7_2)
    (n_6_4) edge [posedge]       node [postxt] 	{\tiny 1}  (n_7_3)
    % Layer 6 neg
    (n_6_3) edge [negedge]       node [negtxt] 	{\tiny -1}  (n_7_2)
    ;

    \draw[dashed] (0+2.05,0.5) rectangle (5.95,-3.135);
    \node[layernode] at (4,-2.9) {$L_1$};

    \draw[dashed] (4+2.05,0.5) rectangle (9.95,-3.135);
    \node[layernode] at (8,-2.9) {$L_2$};

    \draw[dashed] (8+2.05,0.5) rectangle (13.95,-3.135);
    \node[layernode] at (12,-2.9) {$L_3$};

  \end{tikzpicture}

  \caption{Example of a neural network with 72 parameters, capable of
    sorting three numbers, constructed using the sorting network from
    \autoref{fig:three_sort_net}. Weights with a value of $0.0$ are
    omitted for clarity.\label{fig:three_sort_nn} }
\end{figure}
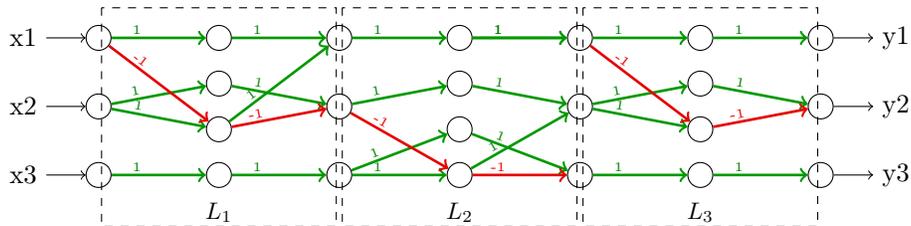

A sorting network can trivially be used to create a neural network for
sorting numbers. For each layer of the sorting network, two hidden
layers in the neural network of appropriate width are created. The
wires are implemented by passing values through neurons with a weight
of 1 (i.e. the values are unchanged). A comparator in the sorting
network is replaced by a comparator neural network after which the two
numbers passed through will be sorted. \autoref{fig:three_sort_net}
shows a minimal sorting network to sort three numbers and
\autoref{fig:three_sort_nn} the corresponding neural network.

As we have shown above, one comparator, implemented using a neural
network, needs three layers. However, since the input and output layer
of comparators following each other can be combined, we need two fully
connected layers in our neural network for each layer in the sorting
network. The first layer needs \(\lceil 1.5x \rceil\) neurons and the second
needs \(x\) neurons to implement one layer of a sorting network which
sorts \(x\) numbers. The number of parameters \(p\) needed for such a
neural network, that implements a sorting network with depth \(d\),
sorting \(x\) numbers are:
\begin{equation}
  p = 2d\lceil 1.5x \rceil x
  \label{eq:param}
\end{equation}
Slightly fewer parameters can be achieved, depending on the exact
structure of the sorting network. More specifically, if a layer of the
sorting network passes through values without processing (i.e. in
cases where a line is not connected to any comparator), some neurons
can be saved. See \autoref{fig:three_sort_nn} as an example: One
neuron can be saved for each second neural network layer because every
layer of the sorting network passes though one value unchanged. If we
assume that the neural sorting network has to be learned from data, we
can not assume an architecture that is so closely fitted to the
problem, and the parameter count of \autoref{eq:param} is more
realistic than the version with neurons removed from individual
layers.

Experimental evidence suggests that neural networks constructed in
this manner are minimal solutions to the sorting problem when using
fully connected layers with rectified linear units as activation
functions and the number of parameters from \autoref{eq:param} is
close to the minimum. The smallest network we could train to sort
three numbers needs 97 parameters, which is higher than our proposed
lower bound from \autoref{eq:param} of 90 parameters.

\subsection{Predicted Number of Parameters for the Identity Task}
\label{sec:pred-numb-param}

Using a typical input image size of \(224 \times 224\) pixels, the amount
of numbers to process depends, as previously mentioned, on whether
there is some form of attentional mechanism or not. \emph{With
  attention}, the system will be able to detect which part of the
image is a patch and which one is not. Therefore, we have to be able
to process the maximal amount of patches that could fit into an image
without overlapping. Assuming a patch size of \(8\times8\), we need to
process a maximum of 784 patches, and therefore numbers. \emph{Without
  attention}, the system has to be able to process all 46,656 possible
patch positions.

Assuming the smaller, information-theoretic, lower bound on the number
of layers needed in a sorting network, we need ten layers and 16
layers with and without attention respectively. Following
\autoref{eq:param}, the neural sorting networks need
\(\approx 18.4 \text{ million}\) parameters with and
\(\approx 65.3 \text{ billion}\) parameters without an attention mechanism.
This difference in parameters shows how important attention is in such
cases to prevent a combinatorial explosion. This need for attention,
to bring the complexity of vision tasks down to practical levels, was
already shown by \cite{tsotsos1988complexity}.

\begin{figure}[tbp]
  \centering
  \begin{tikzpicture}[rotate=-90]
    \useasboundingbox(0,0) rectangle (4, 6.5);
    \usetikzlibrary{shapes.misc, positioning}
    \tikzstyle{startnode}=[left, scale=1]
    \tikzstyle{wire}=[above right, scale=0.6]
    \tikzstyle{innernode}=[above right, scale=0.7]
    \tikzstyle{endnode}=[right, scale=1]
    \tikzstyle{layernode}=[scale=0.9]

    \draw[dashed] (0.5,0) rectangle (1.5,6.5);
    \node[layernode] at (1,0.3) {$L_1$};
    \draw[dashed] (2,0) rectangle (3,6.5);
    \node[layernode] at (2.5,0.3) {$L_2$};

    \foreach \a in {1,...,6}
    \draw[thick, >->] (0,\a) -- ++(3.5,0);
    \foreach \x in {{1,6},{1,5},{1,4},{1,3},{1,2},{1,1},{2.5,5},{2.5,4},{2.5,3},{2.5,2}}
    \filldraw (\x) circle (2pt);

    \draw[thick] (1,6) -- (1,5);
    \draw[thick] (1,4) -- (1,3);
    \draw[thick] (1,2) -- (1,1);
    \draw[thick] (2.5,5) -- (2.5,4);
    \draw[thick] (2.5,3) -- (2.5,2);

    \node[startnode] at (0,4) {};
    \node[startnode] at (0,3) {};
    \node[startnode] at (0,2) {};
    \node[startnode] at (0,1) {};

    \node[innernode] at (1,4) {};
    \node[innernode] at (1,3) {};
    \node[innernode] at (1,2) {};
    \node[innernode] at (1,1) {};

    \node[innernode] at (2,4) {};
    \node[innernode] at (2,3) {};
    \node[innernode] at (2,2) {};
    \node[innernode] at (2,1) {};

    \node[innernode] at (4,4) {};
    \node[innernode] at (4,3) {};
    \node[innernode] at (4,2) {};
    \node[innernode] at (4,1) {};

    \node[innernode] at (6,4) {};
    \node[innernode] at (6,3) {};
    \node[innernode] at (6,2) {};
    \node[innernode] at (6,1) {};
  \end{tikzpicture}
  \caption{A sorting network that is able to sort a list of six
    numbers by repeated application.\label{fig:bubble_sort} }
\end{figure}
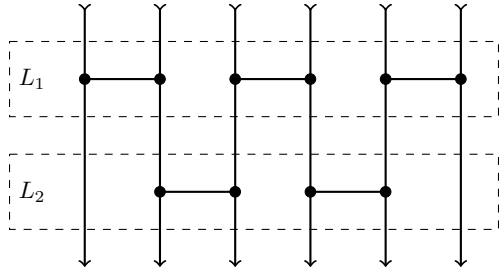

\subsection{Advantages of Iterative Processing}
\label{sec:iterative-processing}

We have previously mentioned, that we suspect that iterative
processing should, similarly to attention, considerably reduce the
complexity of the identity task. This advantage of iterative
processing is connected to the problem that purely feed forward
networks have with sorting numbers.

If we are allowed to send a list of numbers through a sorting network
and apply the same sorting network to the resulting list of numbers
repeatedly, we can sort any list of numbers with a network with only
two layers. Such a network can be implemented by connecting every
second pair of wires with a comparator, starting with the first wire
in layer one, and starting with the second wire for layer two (see
\autoref{fig:bubble_sort} for such a sorting network for six numbers).

Sending numbers through this sorting network ensures that each number
will move at least one step in the direction of its correctly sorted
position. Therefore, a list of \(x\) numbers will be completely sorted
after applying this sorting network \(x\) times.

We can convert this sorting network into a neural network, using the
procedure form Section \ref{sec:orgbc7cb30}. Using \autoref{eq:param},
we can calculate that a neural network that can be used to sort \(x\)
numbers using iterative (recurrent) processing, needs
\(4\lceil 1.5x \rceil x\) parameters. This reduces the network size for our
identity task to \(\approx 7.3 \text{ million}\) parameters with and
\(\approx 26.1 \text{ billion}\) parameters without an attention mechanism.

Suppose we also allow for iterative processing of the list itself. In
that case, the sorting problem can be solved by iteratively applying
the same comparator on each pair of numbers of the list. This approach
is in effect an implementation of bubble sort and can be solved with a
single neural network with only 12 parameters.

This reduction in parameter counts does not only reduce the amount of
resources such neural networks need, it also drastically reduces the
amount of training data needed and leads to better generalizability of
the networks.

In Section \ref{sec:sorting}, we will show experimentally how useful
iterative (recurrent) processing is for our proposed problem.

\section{Experimental Evaluation \label{sec:experiments}}
\label{sec:orgc7b006c}
In the following sections, we will substantiate our theoretical
findings with experimental results. We will show that detecting
identical numbers in a list of numbers (which is, as we showed, the
most efficient input the convolutional part of a CNN can give to the
fully connected part for the identity task) becomes much easier, once
the list of numbers is sorted. We will also show that recurrent
networks perform much better at this task. Besides, we will highlight
that an attentional mechanism makes the identity task much easier to
solve.
\subsection{The Importance Of Sorting}
\label{sec:sorting}
As we have shown in Section \ref{sec:application}, the identity task
boils down to the decision if at least half of the numbers of a list
of numbers are identical.

To show that sorting this list decreases the difficulty of the problem
in practice, we ran experiments where we used fully connected neural
networks to solve this classification problem. The list of numbers was
either sorted or unsorted before being given to the neural network for
classification. Different architectures of up to ten layers with 100
neurons in each layer were tested for lists of ten numbers. The
dataset consisted of 4,000 randomly generated lists for training and
1,000 for testing. Binary cross entropy was used as a loss, the
networks were optimized using Ranger by \cite{ranger} and ReLU by
\cite{hahnloser2000digital} was used as an activation function for all
but the last layer (which used a logistic sigmoid function). Weights
were initialized following the method proposed by
\cite{he2015delving}.

Since we were not able to get an accuracy above \(0.8\) with the
unsorted list, we also tried to solve this task using \(40,000\)
training samples. Looking at the results in \autoref{tab:sort}, it is
easy to see that the task is much easier, once the list is sorted.
Even with one hidden layer and only ten neurons, we were able to
achieve almost perfect accuracy for sorted lists. On the other hand,
for the unsorted list, only a network with ten layers and 50 neurons
for each layer and a training set of 50,000 samples was able to
achieve an accuracy above \(0.9\). These results show that once the
list of numbers is sorted, the task itself becomes very easy.

\begin{table}[tbp]
  \caption{Accuracy of different fully connected neural network
    architectures, when trying to classify whether a list of ten
    numbers contains the same number at least five times. The list is
    sorted or unsorted and either a small or large training set is
    provided. Many more combinations were tested and only the more
    interesting combinations are shown. \label{tab:sort}}
  \centering
  \begin{tabular}{cccccc}
    \toprule
    Input & Hidden & Layersize & Number of & Test & Train\\
          & Layers &           & Parameters & Accuracy  & Accuracy\\
    \midrule
    sorted small & 1 & 10 & 132 & 0.99 & 0.99\\
    sorted small & 1 & 100 & 1302& 1.0 & 1.0\\
    \midrule
    unsorted small & 1 & 500 & 6502 & 0.77 & 0.98\\
    unsorted small & 1 & 1000 & 13002 & 0.80 & 1.0\\
    unsorted small & 2 & 100 & 11402 & 0.79 & 1.0\\
    unsorted small & 2 & 500 & 257002 & 0.80 & 1.0\\
    unsorted small & 2 & 1000 & 1014002 & 0.79 & 1.0\\
    unsorted small & 10 & 100 & 92202 & 0.75 & 1.0\\
    \midrule
    unsorted large & 10 & 5 & 337 & 0.76 & 0.77\\
    unsorted large & 10 & 10 & 1122 & 0.84 & 0.85\\
    unsorted large & 10 & 50 & 23602 & 0.91 & 0.94\\
    \bottomrule
  \end{tabular}
\end{table}

As previously mentioned, we hypothesize that iterative processing
would solve many of the shortcomings of neural networks for these
kinds of problems. To test this hypothesis, we also tested the
previous dataset with a Long Short Term Memory (LSTM) architecture by
\cite{hochreiter1997long}, which contains recurrent connections and
can iteratively process data because of this. The results of this
architecture for unsorted lists can be seen in \autoref{tab:lstm}. It
is evident that a recurrent architecture has two advantages when
solving this kind of problem: First, the networks need far fewer
parameters and second, they can solve the problem using less training
data because they generalize much better, which is a consequence of
them being able to solve the problem with fewer parameters.

\begin{table}[tbp]
  \centering
  \caption{Accuracy of different LSTM neural network architectures,
    when trying to classify whether a list of ten numbers contains the
    same number at least five times. The lists were unsorted.}
  \begin{tabular}{cccccc}
    % BEGIN RECEIVE ORGTBL lstm
\toprule
Training & Layers & Hidded State & Number of & Train & Test\\
Samples &  & Size & Parameters & Accuracy & Accuracy\\
\midrule
4000 & 1 & 5 & 152 & 0.84 & 0.81\\
4000 & 1 & 10 & 502 & 0.91 & 0.88\\
4000 & 1 & 15 & 1052 & 0.92 & 0.88\\
4000 & 2 & 5 & 374 & 0.87 & 0.87\\
4000 & 2 & 10 & 1342 & 0.91 & 0.89\\
4000 & 2 & 15 & 2912 & 0.95 & 0.92\\
\midrule
40000 & 1 & 5 & 152 & 0.86 & 0.86\\
40000 & 1 & 10 & 502 & 0.93 & 0.93\\
40000 & 1 & 15 & 1052 & 0.95 & 0.95\\
40000 & 2 & 5 & 374 & 0.87 & 0.87\\
40000 & 2 & 10 & 1342 & 0.95 & 0.95\\
40000 & 2 & 15 & 2912 & 0.96 & 0.95\\
\bottomrule
    % END RECEIVE ORGTBL lstm
  \end{tabular}
  \label{tab:lstm}
\end{table}

\begin{comment}
  #+ORGTBL: SEND lstm orgtbl-to-latex :booktabs t :splice t :skip 0
  | Training | Layers | Hidded State |  Number of |    Train |     Test |
  |  Samples |        |         Size | Parameters | Accuracy | Accuracy |
  |----------+--------+--------------+------------+----------+----------|
  |     4000 |      1 |            5 |        152 |     0.84 |     0.81 |
  |     4000 |      1 |           10 |        502 |     0.91 |     0.88 |
  |     4000 |      1 |           15 |       1052 |     0.92 |     0.88 |
  |     4000 |      2 |            5 |        374 |     0.87 |     0.87 |
  |     4000 |      2 |           10 |       1342 |     0.91 |     0.89 |
  |     4000 |      2 |           15 |       2912 |     0.95 |     0.92 |
  |----------+--------+--------------+------------+----------+----------|
  |    40000 |      1 |            5 |        152 |     0.86 |     0.86 |
  |    40000 |      1 |           10 |        502 |     0.93 |     0.93 |
  |    40000 |      1 |           15 |       1052 |     0.95 |     0.95 |
  |    40000 |      2 |            5 |        374 |     0.87 |     0.87 |
  |    40000 |      2 |           10 |       1342 |     0.95 |     0.95 |
  |    40000 |      2 |           15 |       2912 |     0.96 |     0.95 |
\end{comment}

\subsection{Training Neural Sorting Networks from Data}
In Section \ref{sec:orgbc7cb30}, we have shown how sorting networks
can be used to construct neural networks that can sort numbers. Since
we expect our networks to learn the sorting operation as part of a
more extensive network, we will experimentally determine how many
parameters neural networks need to be able to learn the sorting
operation from data alone.

As training data, the networks were provided with an unlimited supply
of unsorted vectors containing numbers from \(0\) to \(1\) as input
and the correctly sorted vectors as training targets. The networks
consisted of a variable amount of fully connected layers, all
containing the same amount of neurons (except for the input and output
layers). The networks were again optimized using Ranger by
\cite{ranger}, using ReLU by \cite{hahnloser2000digital} as the
activation function. Weights were initialized following the method by
\cite{he2015delving}, and we used mean squared error as the loss
function.

The number of layers and neurons per layer were systematically tested
multiple times. We consider a network to have learned the sorting task
if it reaches a loss smaller than $1\cdot 10^{-5}$. We report the
networks with the lowest number of parameters. Unfortunately, we were
not able to meaningfully search for networks that sort more than five
numbers. Sorting a list of numbers is in itself a surprisingly tricky
problem to learn for networks and finding a solution is very sensitive
to the weight initialization. Even with the reported smallest
networks, we had to test the same architecture more than 50 times to
be able to teach the network to sort, and this problem became
exponentially more difficult with each new number. Thus, we were not
able to perform a meaningful search for the smallest networks sorting
six numbers and above.

\autoref{tab:sortnetworks} shows the smallest networks we could find
through training. The configurations in the table show the number of
neurons per layer. As can be seen, the number of parameters of trained
networks is strictly greater than the number of parameters needed by
networks that are constructed by the method presented in Section
\ref{sec:orgbc7cb30} and the difference grows rapidly with the size of
the vector to be sorted.

\begin{table}[tbp]
  \centering
  \caption{Smallest neural networks for sorting that could be found by
    training, in comparison to the smallest networks that can be
    constructed using the method from Section \ref{sec:orgbc7cb30}.}
  \begin{tabular}{cccc}
    % BEGIN RECEIVE ORGTBL sortnetworks
\toprule
Numbers & Learned Layer & Parameters & Parameters\\
to Sort & Configurations & Learned & Constructed\\
\midrule
3 & 3 7 6 3 & 97 & 72\\
4 & 4 7 7 7 7 4 & 235 & 144\\
5 & 5 11 11 11 11 11 11 11 11 11 11 11 5 & 1446 & 340\\
\bottomrule
    % END RECEIVE ORGTBL sortnetworks
  \end{tabular}
  \label{tab:sortnetworks}
\end{table}

\begin{comment}
  #+ORGTBL: SEND sortnetworks orgtbl-to-latex :booktabs t :splice t :skip 0
  | Numbers | Learned Layer                        | Parameters |  Parameters |
  | to Sort | Configurations                       |    Learned | Constructed |
  |---------+--------------------------------------+------------+-------------|
  |       3 | 3 7 6 3                              |         97 |          72 |
  |       4 | 4 7 7 7 7 4                          |        235 |         144 |
  |       5 | 5 11 11 11 11 11 11 11 11 11 11 11 5 |       1446 |         340 |
\end{comment}

\subsection{The Influence of Attention \label{sec:attention}}
To show that attention is useful for solving the identity task, we
trained a ResNet-18 architecture by \cite{he2016deep} (which was
already pre-trained on the ImageNet dataset) for two different
variants of the task. In both cases, all images contain three patches
with a size of \(3\times3\). The images were generated at a resolution of
\(14\times14\) and scaled up afterwards to the typical resolution for
ResNet of \(224\times224\). In addition, the background of the image is
kept grey, to make it easier for the system to detect which part
belongs to a patch and which does not. The images can be classified in
the two classes ``identity'' if two of the three patches are identical
and ``non-identity'' if they are not.

\begin{figure}[tbp]
  \centering
  \subfloat[Without attention (one channel)]
  {\includegraphics[width=0.20\textwidth]{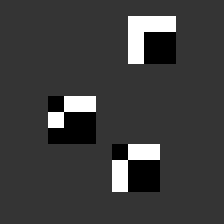}}
  \hspace{4mm}
  \subfloat[With attention (three channels)]
  {\includegraphics[width=0.20\textwidth]{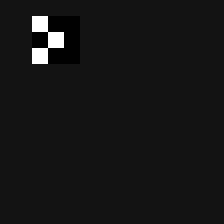}
    \includegraphics[width=0.20\textwidth]{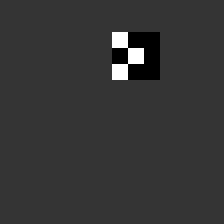}
    \includegraphics[width=0.20\textwidth]{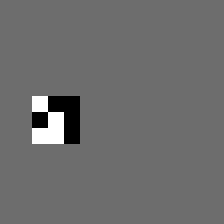}
  }
  \hfill
  \caption{Example images for the two tested versions of the identity
    task. In a) all patches are combined into one channel. In b) the
    patches are presented to the system in different channels, which
    simulates pre-attention. In this case, a) is an example of the
    non-identity-class and b) of the identity class. The varying
    darkness of the images stems from the pre-processing for ResNet.}
  \label{fig:threepatches}
\end{figure}

For the \emph{first variant}, the three patches are presented as a
regular image to the neural network. The three patches are presented
to the network in three separate channels in the \emph{second
  variant}. See \autoref{fig:threepatches} for example images of those
two variants.

The separation for the second variant, in essence, pre-attends the
input, an observation that \cite{kim2018not} already made. The authors
were also able to show that separating objects into separate channels
vastly improves performance on the SVRT dataset by
\cite{fleuret2011comparing} and make the task very easy to solve for
CNNs.

For our experiments, the amount of training data that was available to
the system was not limited, and no care was taken to separate training
from testing data. Although the task looks easy on the surface, the
neural network seems to solve it mainly by memorization, since even
restricting the training data to 50.000 images leads to overfitting on
the training data from the start and an accuracy on the validation set
of about 0.6 (presumably because of some overlap of the training and
validation set).

\begin{figure}[tbp]
  \centering
  \includegraphics[width=0.9\textwidth]{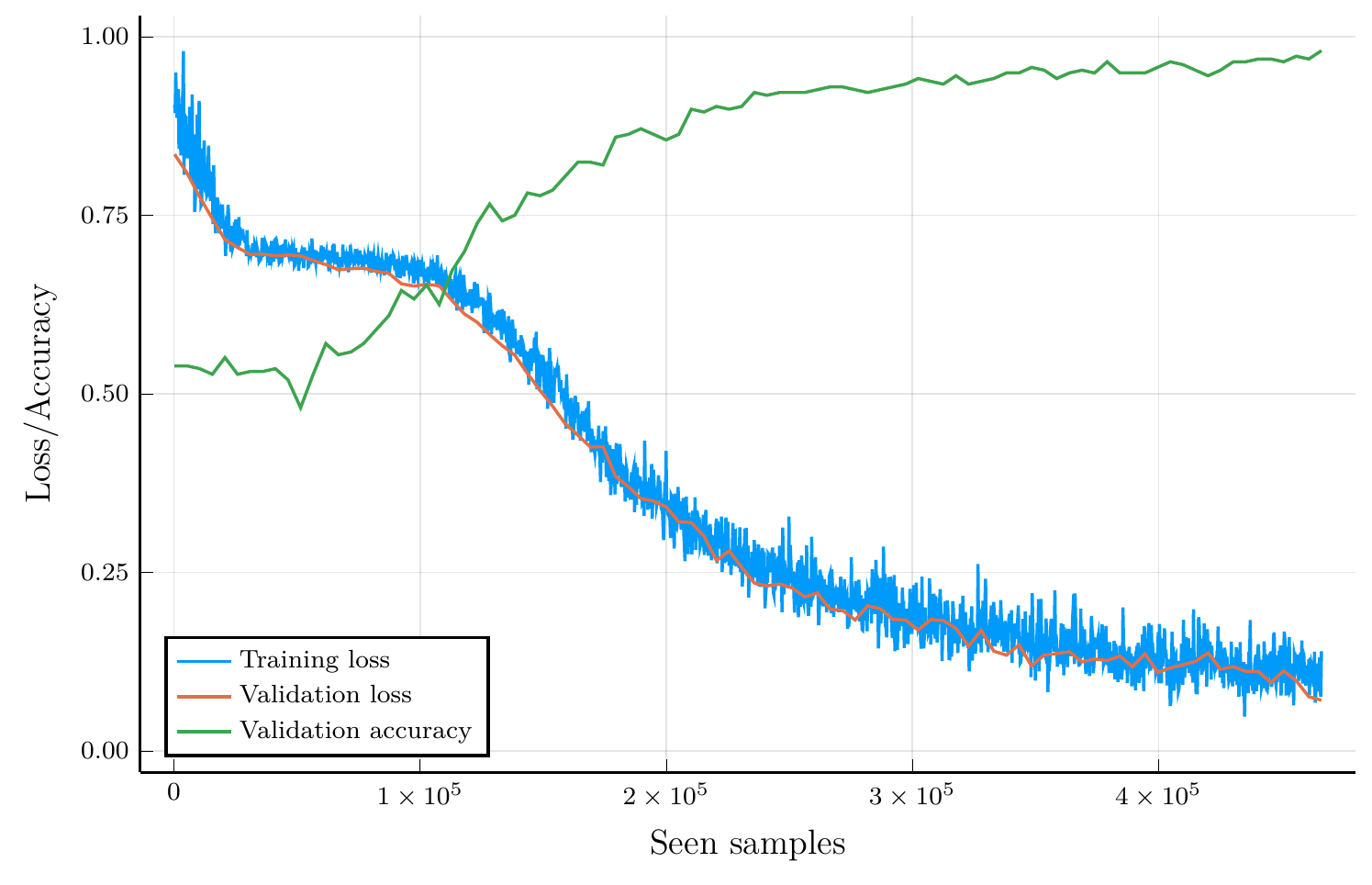}
  \caption{Loss and accuracy during training of a ResNet-18, trained
    on data \emph{with} pre-attention.\label{fig:attention}}
\end{figure}

\begin{figure}[tbp]
    \centering
    \includegraphics[width=0.9\textwidth]{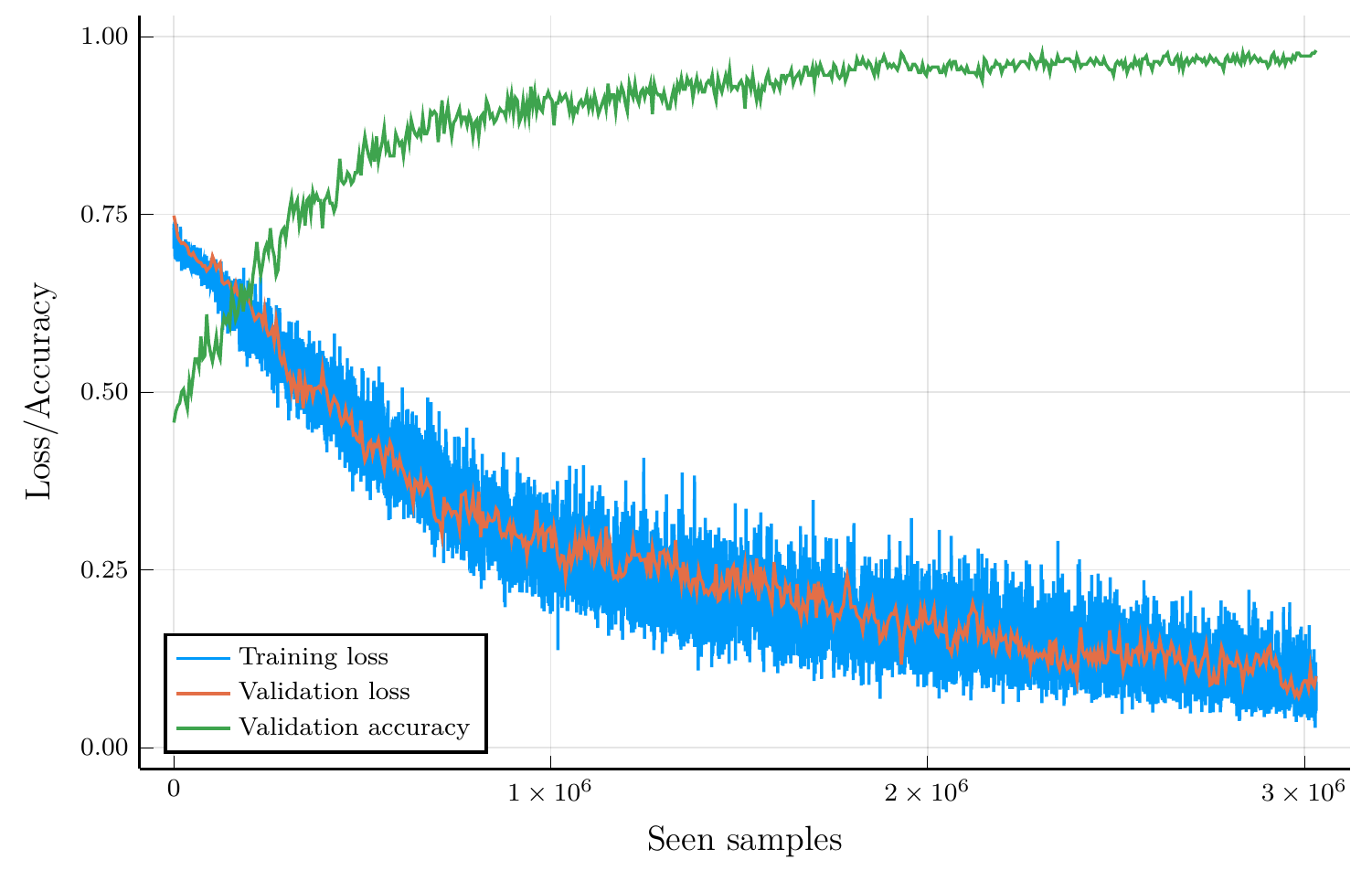}
    \caption{Loss and accuracy during training of a ResNet-18, trained
      on data \emph{without} pre-attention.\label{fig:noattention}}
\end{figure}

ResNet-18 was trained on both variants until an accuracy of 0.98 was
achieved on the validation set. Graphs of the losses and validation
accuracy can be seen in Figures \ref{fig:noattention} and
\ref{fig:attention}.

Comparing the results with and without pre-attention shows that the
task becomes much easier when pre-attention is used. With attention,
around 500,000 images are needed for an accuracy above 0.98, while the
network without attention needs around 3,000,000 images, which amounts
to six times more training data and training time.

\section{Conclusions and Discussion \label{sec:conclusion}}
As we have shown, the reason that CNNs perform poorly on the identity
task is threefold:

\emph{First}, the convolutional layers cannot perform much meaningful
work, since the identity task is inherently global, but convolutions
are inherently local. This mismatch leaves most of the work for the
fully connected part of the network, which operates on a global level.

\emph{Second}, since CNNs usually do not provide an effective
attention mechanism, the number of features forwarded to the fully
connected layer is unnecessarily high. At the level of the fully
connected layers, the identity task can mostly be reduced to a sorting
problem. We used sorting networks to propose a lower bound on the
number of parameters a neural network needs to solve this task.

\emph{Third}, since most of the processing has to be done in fully
connected layers which do not share weights, without recurrent
connection or other forms of iterative processing, the resulting
system is also using data very inefficiently. This inefficiency can be
seen by the fact that even only looking at the fully connected part of
the problem, solving the identity task for ten numbers needs 50,000
training samples to solve.

All those arguments do not necessarily mean that smaller networks can
not solve the identity task, but they likely will not generalize well
and only solve the problem with memorization.

In our opinion, two mechanisms are needed to solve comparison tasks
efficiently (regarding the number of parameters of the networks as
well as the amount of training data needed). First, an attention
mechanism to reduce the number of entities that need to be compared to
prevent a combinatorial explosion is needed. Second, the information
extracted from these entities has to be processed iteratively using a
recurrent architecture to reduce the number of parameters, as well as
the amount of training data needed through weight sharing.

% \section{ToDo}
% \david{We could write this as hypothesis which is evaluated in the
%   experimental section. E.g.: \textbf{Hypothesis 2: } If no
%   attentional mechanism is present the complexity to solve non-local
%   tasks increases exponentially.} \sebastian{Could not do that at the
%   original position, but probably worth to see if we can put it
%   somewhere.}

\section{Bibliography}
\label{sec:org2d3a25c}
\bibliography{myref}
\end{document}